\newif\ifarchive
\archivetrue

\documentclass[%
reprint,
amsmath,amssymb,
aps,
prx,
]{revtex4-1}

\usepackage{tikz}

\usepackage{amsmath, amsthm, amsfonts}

\theoremstyle{plain}
\newtheorem{thrm}{Theorem}[section] 
\newtheorem{prop}[thrm]{Proposition}

\theoremstyle{definition}
\newtheorem{defn}[thrm]{Definition}
\newtheorem{axiom}[thrm]{Axiom}
\newtheorem*{principle*}{Principle of scientific objectivity}
\theoremstyle{remark}

\usepackage[only,llbracket, rrbracket,llparenthesis,rrparenthesis]{stmaryrd}

\usepackage{enumitem}
\setlist{noitemsep}

\usepackage{MnSymbol}

\DeclareMathOperator{\truth}{truth}
\DeclareMathOperator{\possFn}{poss}

\def\TRUE{\textsc{true}}
\def\FALSE{\textsc{false}}

\def\stmtSet{\mathcal{S}}
\def\vstmtSet{\mathcal{S}_\textsf{v}}

\def\tautology{\top}
\def\contradiction{\bot}

\def\comp{\doublefrown}
\def\ncomp{\ndoublefrown}

\def\narrower{\preccurlyeq}
\def\nnarrower{\npreccurlyeq}
\def\broader{\succcurlyeq}
\def\nbroader{\nsucccurlyeq}

\def\indep{\upmodels}
\def\nindep{\nupmodels}

\def\AND{\wedge}
\def\bigAND{\bigwedge}
\def\OR{\vee}
\def\bigOR{\bigvee}
\def\NOT{\neg}

\newcommand{\stmt}[1][s] {\mathsf{#1}}

\newcommand{\obs}[1][s] {\mathsf{#1}}

\newcommand{\edomain}[1][D] {\mathcal{#1}}

\newcommand{\tdomain}[1][D] {\bar{\mathcal{#1}}}

\newcommand{\basis}[1][B] {\mathcal{#1}} 

\newcommand{\statement}[1] {\emph{``#1''}}

\begin{document}

\title{Towards a general mathematical theory of experimental science}

\author{Gabriele Carcassi}
\email{carcassi@umich.edu}
\author{Christine A. Aidala}
\affiliation{
	Physics Department \\
	University of Michigan \\
	Ann Arbor, MI 48109 \\
}

\date{\today}

\begin{abstract}
	We lay the groundwork for a formal framework that studies scientific theories and can serve as a unified foundation for the different theories within physics. We define a scientific theory as a set of verifiable statements, assertions that can be shown to be true with an experimental test in finite time. By studying the algebra of such objects, we show that verifiability already provides severe constraints. In particular, it requires that a set of physically distinguishable cases is naturally equipped with the mathematical structures (i.e. second-countable Kolmogorov topologies and $\sigma$-algebras) that form the foundation of manifold theory, differential geometry, measure theory, probability theory and all the major branches of mathematics currently used in physics. This gives a clear physical meaning to those mathematical structures and provides a strong justification for their use in science. Most importantly it provides a formal framework to incorporate additional assumptions and constrain the search space for new physical theories.
\end{abstract}

\maketitle


\section{Introduction}

When considering physics as a discipline, one cannot help but notice that it is essentially composed of different theories and models loosely connected to each other, each with its own starting points and realm of applicability. Classical mechanics, electromagnetism, general relativity, quantum mechanics, thermodynamics and so on may share some general ideas but, in the end, they all have their own separate foundation, either in a different set of laws and principles or simply in positing a unique mathematical structure. This state of affairs is so entrenched in our field it somehow feels like the proper, if not the only, approach. But is it? Or does this approach actually hinder progress and understanding?

We can gain perspective by comparing to different fields of science and engineering. For example, in computer science theory one starts by defining symbolic languages and computational devices~\cite{Turing} and then shows that no algorithm exists that can correctly decide whether an arbitrary program terminates given an arbitrary input~\cite{Sipser}. In communication theory one defines channels and the information they carry~\cite{Shannon} and proves general results such as calculating the maximum rate at which information can be transmitted over a channel given a set amount of noise~\cite{Pierce}. In control theory one defines a system in terms of state, inputs and outputs~\cite{Brogan} and then looks for general strategies for control such as Kalman filtering~\cite{Kalman}. These theories, with their respective mathematical structures and results, are \emph{general} in the sense that they apply to all control systems, all communication systems and all computational devices. More specific topics in each field are constructed by further constraining those general mathematical structures. In other words, the subject matter has been properly defined formally and therefore the results follow simply from the mere definition of the problem.

In physics there is no equivalent. Classical mechanics, thermodynamics and quantum mechanics are not specific topics within some more general formal theory. But how would that work? As in the other general theories, we need to clarify what our starting points are and let the logic follow. The first requirement of a physical theory, or rather of any scientific theory in general, is that it is experimentally testable. So we would start by characterizing the properties and limitations of experimental verification. The second requirement, which is where physics starts, is identifying states and processes. After those are characterized in general, we can then specialize them for different cases. Some processes will be deterministic and reversible. Some processes will be non-deterministic and some will have equilibria. Some states can be broken into parts and some cannot. By adding different assumptions on states and processes we would recover classical Hamiltonian mechanics in one case, quantum mechanics in another and thermodynamics in yet another. In all these theories, some basic properties will be common simply because states need to be identified experimentally, and some properties will be different because the type of state or the type of process is different. This is what we mean by a general mathematical theory of experimental science.


\begin{figure}\label{twoQuestions}
	\centering
	\begin{tikzpicture}
	[child anchor = north, every node/.style={draw, align = center, inner sep=5pt},
	edge from parent/.style={draw, ->, -latex, stealth-},
	level 1/.style={level distance=1cm},
	level 2/.style={level distance=1.3cm, sibling distance=2.4cm}, sloped]
	\node {Experimental verifiability}
	child {node {States and processes}
		child {
			node {Classical \\ mechanics}
		}
		child {
			node {Quantum \\ mechanics}
		}
		child {
			node {Thermo- \\dynamics}
		}
		child {
			node {...}
		}
	};
	\end{tikzpicture}
	\caption{Overall structure for a general mathematical theory of experimental science. The starting point is experimental verification. States and processes are particular types of experimentally testable objects. Different theories in physics describe particular states under particular processes.}
\end{figure}
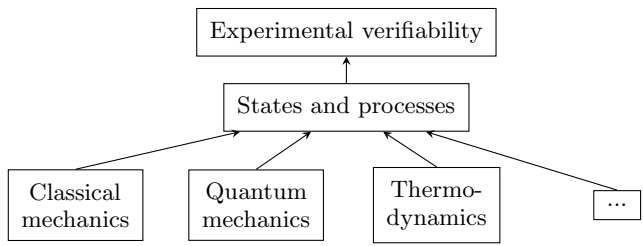

Such a general theory would force us to clarify our assumptions, thus the realm of applicability of each theory. It would always put physics at the center of our discussion, as physical ideas are the starting point for our formal framework and not after-the-fact interpretations. It would give science sturdier mathematical grounds, as each mathematical symbol is given a precise meaning and non-physical objects are excluded by construction. It would foster connections between different fields of knowledge: nature does not care about such divisions. It would provide a framework in which to pose new questions and solve old ones. In other words: it would provide a better foundation for physics and for the rest of the sciences. Developing a general mathematical theory for experimental science could be compared to what happened in mathematics during the first half of the last century, when it was reorganized using logic and set theory as its foundation, which had profound repercussions in the field. Our preliminary work in both physics~\cite{Carc1} and math~\cite{Carc2} convinced us that such a goal is possible and within reach.

The aim of the present work is to lay down the beginning of a general mathematical theory of experimental science, limited to the part concerning experimental verification. We start by defining verifiable statements: assertions whose truth can be verified experimentally in finite time. We study the logic of these statements, which is different from the standard Boolean logic because of finite time termination. We define experimental domains as a collection of verifiable statements that, if given an indefinite amount of time, we can keep testing, forever refining our knowledge. From each experimental domain we construct its possibilities: the different cases we can experimentally distinguish. The main result is that \textbf{each set of experimentally distinguishable possibilities comes equipped with a natural second-countable Kolmogorov topology, where each open set corresponds to a verifiable statement, and a natural $\sigma$-algebra, where each Borel set corresponds to a theoretical statement which gives predictions for verifiable ones.} For example, \statement{the mass of the particle is more than $0.4$ and less than $0.6$ MeV} is a verifiable statement precisely because $(0.4, 0.6)$ is an open set, while \statement{the mass of the particle is exactly $0.5$ MeV} is not verifiable because $[0.5,0.5]$ is not an open set.\footnote{The standard topology of the real numbers keeps track of finite precision measurements.} From that general result we can derive the following conclusion: \textbf{any set of physically distinguishable cases has at most cardinality of the continuum}. That is, the only thing we can do to distinguish a particular case is run the tests for a countable set of statements, the output of which can be imagined as a countable sequence of true and false. This is equivalent to the binary representation of a real number. In a nutshell, \textbf{experimental verification by itself guarantees us the existence of two mathematical structures that are the foundations of most tools used in physics}, such as differential geometry (Riemannian and symplectic), measure theory, probability theory and many others.


While this result is remarkable by itself, the most interesting aspect is that it can be used as a cornerstone for a more general theory. Conceptually, in the same way that topological spaces keep track of what can be verified experimentally, manifolds keep track of objects that can be identified by continuous quantities, differentiable manifolds keep track of objects that allow a density to be defined over them and symplectic manifolds (i.e. phase space of classical mechanics) allow the density to be coordinate invariant (i.e. observer independent). By doing all this work formally we will be forced to recognize the assumptions that go into those structures, and understand if and how they fail. For example, we are currently in the process of identifying the necessary and sufficient conditions such that a set of verifiable statements defines a continuous quantity. While, as one would expect, these can only correspond to an idealized case where one pretends a better precision measurement can always be achieved, they also tell us precisely how this idealized case fails which can be relevant to work on Planck scale physics. But whatever other tools we will have in that case, we \emph{know} we will always have at least a topology and a $\sigma$-algebra.

As another example, note that the set of discontinuous functions from $\mathbb{R}$ to $\mathbb{R}$ has cardinality greater than the continuum and therefore cannot represent a set of physically distinguishable objects. This means that the square integrable space of functions typically used in quantum mechanics is far too big, and, in fact, it contains unphysical elements, such as states with infinite energy. However, the Schwartz space, like the set of continuous functions, has cardinality of the continuum and can be given a second-countable Kolmogorov topology.\footnote{It also has other properties that map to physical requirements. For example, quantities of the form \unexpanded{$\langle \psi|Q^n P^m|\psi\rangle$} are always finite and the Fourier transform is always well defined.} To our understanding, the reason why Hilbert spaces are mathematically useful is because one can take limits and use the vector space norm to converge. However, we may instead be able to use the corresponding $\sigma$-algebra to take limits, like one does in probability and measure theory. In other words, we may be able to construct, within this framework, a more mathematically sound and physically motivated foundation for quantum mechanics that includes the objects and only the objects that are physically meaningful.

As a last example, we also believe we can form a connection to computer science. We can characterize the output of a computational device as a set of verifiable statements (e.g. \statement{the third bit of output is 1}) that are a function of the input, which can also be characterized as verifiable statements (e.g. \statement{the second bit of the input is 0}). In this context, all possible inputs form the possibilities of our space, as they define all possible outputs of the computation. Now suppose we have a function from the input to a Boolean value. Then we can write the statements \statement{the input is such that the output of the function is true} and \statement{the input is such that the output of the function is false}. If both these statements correspond to open sets on the possible inputs then they are verifiable statements and we construct a test (i.e. a program) that will always terminate. That is, we can combine tools from topology and computer science to answer a question of the type: is this scientific problem testable or computable? In a similar way, connections to information theory can also be established: a deterministic and reversible process, in fact, is one for which description of the input is equivalent to the description of the output, which means information entropy is conserved. These links illustrate the potential as a theory of experimental science, and not just physics.

We hope that it is clear by now that a general mathematical theory of experimental science is missing, is possible and would be extremely useful. Whether you work on condensed matter, quantum thermodynamics, complex systems, particle physics, gravitation or grand unified theories, if what you are doing is science you will have, at some point, experimental verification. You will have problems (often the most interesting ones) that arise from questions such as: what is it that I can measure? How do I do it? Under what assumptions do the quantities in my theory map to those measurements? When are they even well defined? What can I compute within my theory? What is testable? Currently, you do not have a general framework to pose those questions. And it turns out you do not need a completely ad-hoc one for your specific field. The mere \emph{logic} of experimental verification requires a particular structure. And there is a lot more to experimental verification than its logic: the quantity itself has to, at least in a sense, exist over a finite period of time to be measurable, there must exist a process to transfer that information to a measurement device, the system cannot be assumed to be always completely isolated as it would not be physically accessible. If you are working on foundational aspects, chances are you may already be thinking about some of these issues as they relate to your field. But what parts of those problems are specific to your field? How much can be treated generally, so that not only each field has less work to do but also we have a common language across fields? What are the assumptions specific to your field and those that are more general? A general mathematical theory of experimental science would provide you with a formal framework to precisely pose and possibly answer those questions. As the starting points are clarified, the foundations of different fields may be affected, or even partially integrated, and potentially change how they are taught as well.

The irony is that while the project is extremely ambitious, the biggest obstacles are not, at this point, technical. The biggest obstacles are sociological. On one side the nature of this work is extremely interdisciplinary. On the other side math and physics are divided into fields and sub-fields that are increasingly narrow.\footnote{We attended a topology conference where there seemed to be a disconnect between what mathematicians and condensed matter physicists meant by topology, to the point that they frequently talked past each other.} This means that this work does not fit naturally within any community while requiring technical expertise from many. To obviate these problems, borrowing practices from the open source software community, we are developing our main body of work in the open~\cite{Carc3}, so that we can address the topic in its entirety, and our colleagues in philosophy, mathematics and physics can all check their respective parts. This article is the product of that process and focuses on current results of greater interest to physicists, omitting more mathematical and philosophical details that can be found in the broader work.

\section{The principle of scientific objectivity}

As the starting point of our general theory, we introduce the following guiding principle:

\begin{principle*}
Science is universal, non-contradictory and evidence based.
\end{principle*}

This means that science concerns itself only to the study of assertions that have a well defined truth value which can be verified experimentally. The issue at hand is to formally capture this informal intuition. We begin with the following common definition.

\begin{defn}
	The \textbf{Boolean domain} is the set $\mathbb{B} = \{\FALSE, \TRUE\}$ of all possible truth values.
\end{defn}

Next we need to define our truth bearer. In mathematical logic, this is typically a well formed formula. This cannot work for us: what we are interested in is the meaning of the assertion and not how it is expressed. For example, \statement{this animal is a dog} and \statement{questo animale \`e un cane} represent the same fact expressed in different languages. As science is universal, it should not matter the language, units or reference system used to make an assertion.\footnote{In the same vein, a statement is not necessarily one sentence as it has to include all the information needed to understand what exactly we are talking about. These are the type of philosophical issues that are discussed at greater length in~\cite{Carc3}.} While we will still use standard mathematics for the formal system, we introduce a variation of algebraic logic to represent our ``informal'' assertions. We start by defining our truth bearer as:

\begin{axiom}\label{ax_statement}
	A \textbf{statement} $\stmt$ is an assertion that is either true or false. Formally, a statement is an element of the set $\mathcal{S}$ of all statements upon which is defined a function $\truth: \mathcal{S} \to \mathbb{B}$ that returns the truth value for each element.
\end{axiom}

Note how the first part of the definition captures the informal meaning of what we are describing, while the second part captures the part that is formalized. This pattern will be present in most of our definitions and it serves to clarify both what is being formalized and how. Therefore, in science, the statement is an assertion while for the math it is just an element in some set.

While in math the truth value is the focus of the logic system, in science it is generally established experimentally. But our scientific model may constrain certain statements or statement combinations to be ruled out. For example, the statements \statement{this animal is a dog} and \statement{this animal is a cat} can both be either true or false, but the statement \statement{this animal is a cat and a dog} can never be true. That is, the role of our logic system is not to keep track of what is true and false, but of what cannot possibly be true and cannot possibly be false. We need to keep track of these relationships so that we can never have inconsistencies or paradoxes.

\begin{defn}
	Given a collection of statements $\{\stmt_i\}^n_{i=1}$, a \textbf{consistent truth assignment} is a collection of truth values $\{t_i\}^n_{i=1}$ such that it is logically consistent to simultaneously suppose that $\truth(\stmt_i) = t_i$ for all $1 \leq i \leq n$. That is, from those assumptions it cannot be proven that $\truth(\stmt_i) \neq t_i$ for any $1 \leq i \leq n$.  This definition generalizes to the case of infinite, possibly uncountable, indexed families.
\end{defn}

Note that the truth assignments are just hypotheticals. They are not actual physical entities. There is only one truth value, the one found experimentally. Logical consistency is defined on the truth assignments and not on the truth values. Therefore we need a way to track whether the meaning of each statement allows it to be true or not.

\begin{axiom}\label{ax_possibilities}
	The \textbf{possibilities} of a statement $\stmt$ are the possible truth values allowed by the content of the statement. Formally, on $\mathcal{S}$ is also defined a function $\possFn: \mathcal{S} \to \{\{\FALSE, \TRUE\},\{\FALSE\},\{\TRUE\}\}$ such that:
	\begin{itemize}
		\item $\truth(\stmt) \in \possFn(\stmt)$ for all $\stmt \in \mathcal{S}$. This remains valid in every consistent truth assignment.
		\item for any collection of statements $\{\stmt_i\}^n_{i=1}$, for any $1 \leq j \leq n$ and for any $t \in \possFn(\stmt_j)$ there exists a consistent truth assignment $\{t_i\}^n_{i=1}$ such that $t_j = t$. This generalizes to the case of infinite, possibly uncountable, indexed families.
	\end{itemize}
\end{axiom}

With this axiom, we can distinguish between statements that can never be true and those that just happen to be true.

\begin{defn}
	A \textbf{tautology} $\tautology$ is a statement that must be true simply because of its content. That is, $\possFn(\tautology) = \{\TRUE\}$.
\end{defn}

\begin{defn}
	A \textbf{contradiction} $\contradiction$ is a statement that must be false simply because of its content. That is, $\possFn(\contradiction) = \{\FALSE\}$.
\end{defn}

We also need to express relationships between the truth of different statements. For example, if we assign true to \statement{this animal is a dog} then we cannot assign true to \statement{this animal is not a dog}. Therefore we introduce the following:

\begin{axiom}\label{ax_functions_of_statement}
	We can always construct a statement whose truth value arbitrarily depends on an arbitrary set of statements. Formally, given an arbitrary truth function $f_{\mathbb{B}} : \mathbb{B}^n \to \mathbb{B}$ there exists a function $f : \mathcal{S}^n \to \mathcal{S}$ such that
	$$\truth(f(\stmt_1, ..., \stmt_n)) = f_{\mathbb{B}}(\truth(\stmt_1), ..., \truth(\stmt_n))$$
	and the same relationship remains valid in every consistent truth assignment. This also holds in the case of infinite, possibly uncountable, arguments.
\end{axiom}

We will use the standard symbols $\NOT$, $\AND$, $\OR$ to indicate the negation (logical NOT), conjunction (logical AND) and disjunction (logical OR). With these three axioms, we can rederive all of the rules of classical logic. First we define our equivalence.

\begin{defn}
	Two statements $\stmt_1$ and $\stmt_2$ are \textbf{equivalent} $\stmt_1 \equiv \stmt_2$ if they must be equally true or false simply because of their content. Formally, $\stmt_1 \equiv \stmt_2$ if and only if $(\stmt_1 \AND \stmt_2) \OR (\NOT\stmt_1 \AND \NOT\stmt_2)$ is a tautology.
\end{defn}

From this definition, one can prove that two statements are equivalent if and only if they have the same truth in all consistent truth assignments. From that one can prove the following two propositions.

\begin{prop}
	Statement equivalence satisfies the following properties:
	\begin{itemize}
		\item reflexivity: $\stmt \equiv \stmt$
		\item symmetry: if $\stmt_1 \equiv \stmt_2$ then $\stmt_2 \equiv \stmt_1$
		\item transitivity: if $\stmt_1 \equiv \stmt_2$ and $\stmt_2 \equiv \stmt_3$ then $\stmt_1 \equiv \stmt_3$
	\end{itemize}
	and is therefore an \textbf{equivalence relationship}.
\end{prop}

\begin{prop}\label{boolean_properties}
	The set of all statements $\mathcal{S}$ satisfies the following properties:
	\begin{itemize}
		\item associativity: $a \OR (b \OR c) \equiv (a \OR b) \OR c$, $a \AND (b \AND c) \equiv (a \AND b) \AND c$
		\item commutativity: $a \OR b \equiv b \OR a$, $a \AND b \equiv b \AND a$
		\item absorption: $a \OR (a \AND b) \equiv a$, $a \AND (a \OR b) \equiv a$
		\item identity: $a \OR \contradiction \equiv a
		$, $a \AND \tautology \equiv a$
		\item distributivity: $a \OR (b \AND c) \equiv (a \OR b) \AND (a \OR c)$, $a \AND (b \OR c) \equiv (a \AND b) \OR (a \AND c)$
		\item complements: $a \OR \NOT a \equiv \tautology$, $a \AND \NOT a \equiv \contradiction$
		\item De Morgan: $\NOT a \OR \NOT b \equiv \NOT (a \AND b)$, $\NOT a \AND \NOT b \equiv \NOT (a \OR b)$
	\end{itemize}
	This, by definition, means $\mathcal{S}$ is a \textbf{Boolean algebra}.
\end{prop}

One can also prove that the Boolean algebra is complete, and therefore it generalizes to the infinite case.

This system not only allows us to use formal classical logic while keeping the statements informal, but is also equipped to capture causal relationships.\footnote{Note that the axioms prevent any form of paradox, simply because the possibilities of a statement are never empty and a consistent truth assignment must exist. Therefore situations like \statement{this statement is not true} are ruled out simply because they cannot satisfy the axioms.} For example, $\stmt_1=$\statement{the thermometer indicator is between 24 C and 25 C} and $\stmt_2=$\statement{the temperature is between 24 C and 25 C} are equivalent because both $\stmt_1 \AND \NOT \stmt_2$ and $\NOT \stmt_1 \AND \stmt_2$ are contradictions (on the assumption that our thermometer is actually working). We can also define other semantic relationships.

\begin{defn}
	Given two statements $\stmt_1$ and $\stmt_2$, we say that:
	\begin{itemize}
		\item $\stmt_1$ \textbf{is narrower than} $\stmt_2$ (noted $\stmt_1 \narrower \stmt_2$) if $\stmt_2$ is true whenever $\stmt_1$ is true simply because of their content. That is, $\stmt_1 \AND \NOT \stmt_2 \equiv \contradiction$.
		\item $\stmt_1$ \textbf{is broader than} $\stmt_2$ (noted $\stmt_1 \broader \stmt_2$) if $\stmt_2 \narrower \stmt_1$.
		\item $\stmt_1$ \textbf{is compatible to} $\stmt_2$ (noted $\stmt_1 \comp \stmt_2$) if their content allows them to be true at the same time. That is, $\stmt_1 \AND \stmt_2 \nequiv \contradiction$.
		
	\end{itemize}
	The negation of these properties will be noted by $\nnarrower$, $\nbroader$ , $\ncomp$ respectively.
\end{defn}
\begin{defn}
	The elements of a set of statements $S \subseteq \mathcal{S}$ are said to be \textbf{independent} (noted $\stmt_1 \indep \stmt_2$ for a set of two) if their content is such that any combination of their possibilities is allowed. That is, $\possFn(f(S)) = f(\bigtimes\limits_{\stmt \in S} \possFn(\stmt))$ for any truth function $f : \mathbb{B}^{|S|} \to \mathbb{B}$. The negation of independence will be noted by $\nindep$.
\end{defn}

For example, given the following statements:
\begin{description}
	\item $\stmt_1=$\statement{that animal is a cat}
	\item $\stmt_2=$\statement{that animal is a mammal}
	\item $\stmt_3=$\statement{that animal is a dog}
	\item $\stmt_4=$\statement{that animal is black}
\end{description}
we have the $\stmt_1$ $\narrower$ $\stmt_2$ (i.e. it is more specific), $\stmt_1$ $\ncomp$ $\stmt_3$ (i.e. they cannot be true at the same time) and $\stmt_1$ $\indep$ $\stmt_4$ (i.e. the truth of one tells us nothing about the truth of the other).

Another interesting result is that statement narrowness imposes a partial order on the set of all statements.

\begin{prop}
	Statement narrowness satisfies the following properties:
	\begin{itemize}
		\item reflexivity: $s \narrower s$
		\item antisymmetry: if $s_1 \narrower s_2$ and  $s_2 \narrower s_1$ then $s_1 \equiv s_2$
		\item transitivity: if $s_1 \narrower s_2$ and $s_2 \narrower s_3$ then $s_1 \narrower s_3$
	\end{itemize}
	and is therefore a \textbf{partial order}.
\end{prop}

As every element in our general theory will be constructed upon statements, these operations are important as they will characterize all those constructions. For example, if we quantify the precision of a set of statements, it will need to be ordered in a way that is compatible with narrowness. If we define statistical independence, it will have to be defined in a way that is compatible with statement independence. In the same way that set theory defines concepts common across all mathematics, the general theory defines basic concepts that are common to all scientific theories.

\section{Verifiable statements and experimental domains}

We have the tools for dealing with assertions that are universal and non-contradictory, now we have to develop the tools for those that are evidence based as well.

\begin{axiom}\label{ax_verifiable_statements}
	A \textbf{verifiable statement} is a statement that can be shown to be true experimentally. Formally, a statement $\stmt$ is verifiable if it is part of the subset $\stmt \in \vstmtSet \subset \stmtSet$ of all verifiable statements.
\end{axiom}

Physically, this means that we have a repeatable procedure that anybody can execute and that always gives the same result. If the statement is true, then this experimental test must terminate successfully. Note that, in general, the test may not terminate if the statement is false. Consider the following:
\begin{enumerate}
	\item find a swan
	\item if it is black terminate successfully
	\item go to step 1
\end{enumerate}
This will terminate if a black swan is found, but it will not terminate if no black swans exist (i.e. absence of evidence is not evidence of absence). Because of non-termination, failure to verify is not verification of the negation. That is, the negation of a verifiable statement is not necessarily a verifiable statement.

Given two statements, though, we can verify their conjunction simply by verifying both: if they are both true, both their tests will terminate and verify the conjunction. But we cannot extend this to an infinite number of statements as we would never terminate. We can also verify the disjunction of two statements: once one test terminates we are done. And because we only need one test to terminate, we can generalize to a countable number of verifiable statements by following this procedure:
\begin{enumerate}
	\item initialize $n$ to 1
	\item for each $i=1..n$
	\begin{enumerate}
		\item run the test for $\stmt_i$ for $n$ seconds
		\item if it terminates successfully then terminate successfully
	\end{enumerate}
	\item increment $n$ and go to step 2
\end{enumerate}
This procedure will run all tests for an arbitrary amount of time. Therefore, if one statement is true, it will make the test terminate successfully, even if all other tests would never terminate. In light of this, we can set the following axioms.

\begin{axiom}\label{ax_verifiable_AND}
	The conjunction of a finite collection of verifiable statements is a verifiable statement. Formally, let $\{\stmt_i\}_{i=1}^{n} \subseteq \vstmtSet$ be a finite collection of verifiable statements. Then the conjunction $\bigAND\limits_{i=1}^{n} \stmt_i \in \vstmtSet$ is a verifiable statement.
\end{axiom}
	\begin{axiom}\label{ax_verifiable_OR}
	The disjunction of a countable collection of verifiable statements is a verifiable statement. Formally, let $\{\stmt_i\}_{i=1}^{\infty} \subseteq \vstmtSet$ be a countable collection of verifiable statements. Then the disjunction $\bigOR\limits_{i=1}^{\infty} \stmt_i \in \vstmtSet$ is a verifiable statement.
\end{axiom}

We could also define decidable statements as those for which falsehood can also be tested experimentally. Table \ref{tab:algebras} compares the different algebras.

\begin{table*}
	\centering
	\begin{tabular}{p{0.14\textwidth} p{0.08\textwidth} p{0.13\textwidth} p{0.22\textwidth} p{0.23\textwidth}}
		Operator & Gate & Statement & Verifiable Statement & Decidable Statement  \\ 
		\hline 
		Negation & NOT & allowed & disallowed & allowed \\ 
		Conjunction & AND & arbitrary  & finite & finite \\ 
		Disjunction & OR & arbitrary  & countable & finite \\ 
	\end{tabular}
	\caption{Comparing algebras of statements.}\label{tab:algebras}
\end{table*}

Now that we can verify statements one by one, we need to define what it means to verify a group of them. In principle, given a set of verifiable statements we can simply start testing them one after the other. However, if we are given a set of uncountable statements then, even if we have an indefinitely long time at our disposal, we will not be able to create a procedure that eventually tests all statements. However, we may not need to actually run all tests for all statements. For example, if we found that $\stmt_1$ is true then there is no need to test any of its disjunctions like $\stmt_1 \OR \stmt_2$.

\begin{defn}
	Given a set $\edomain$ of verifiable statements, $\basis \subseteq \edomain$ is a \textbf{basis} if the truth values of $\basis$ are enough to deduce the truth values of the set. Formally, all elements of $\edomain$ can be generated from $\basis$ using finite conjunction and countable disjunction.
\end{defn}

Therefore it is the size of the basis that matters and not the size of the set. If the basis is countable we can keep going and, if any statement is true, it will eventually be verified experimentally.\footnote{We assume we are given an indefinitely long time because this handles limits and because we have not given constraints for how long or short a successful test can be (i.e. if we ran a test for three days without terminating, it may still successfully complete in three days and one minute).}

\begin{defn}
	An \textbf{experimental domain} $\edomain$ represents all the experimental evidence that can be acquired about a scientific subject in an indefinite amount of time. Formally, it is a set of statements, closed under finite conjunction and countable disjunction, that includes precisely the tautology, the contradiction, and a set of verifiable statements that can be generated from a countable basis.
\end{defn}

These axioms and definitions formally characterize what we mean by evidence based. A scientific theory will be fully defined by a countable set of verifiable statements, the basis for an experimental domain.

\section{Theoretical domains and possibilities}

While verifiable statements define what can be scientifically studied, not all interesting scientific statements are directly verifiable. Consider the two statements \statement{there exists extra-terrestrial life} and \statement{there is no extra-terrestrial life}. We can verify the first if we happen to find signs of life somewhere, but experimentally verifying the second is practically impossible. Yet, the second is still meaningful as a prediction: it predicts that the test for the first will never terminate. That is, while negations are not verifiable we can still logically talk about them when, for example, constructing truth assignments.

\begin{defn}
	The \textbf{theoretical domain} $\tdomain$ of an experimental domain $\edomain$ is the set of statements that we can use to state predictions, which is constructed from $\edomain$ by allowing negation. We call \textbf{theoretical statement} a statement that is part of a theoretical domain. More formally, $\tdomain$ is the set of all statements generated from $\edomain$ using negation, finite conjunction and countable disjunction.
\end{defn}

Note that the new statements provide no new information. In fact, all statements in the theoretical domain $\tdomain$ can be generated by negation, countable conjunction and countable disjunction from a basis $\basis$ of $\edomain$. As they provide all possible predictions, we focus on the ones that completely specify the truth value for all theoretical statements. For example, once we know that \statement{this animal is a cat} we know that \statement{this animal has whiskers}, that \statement{this animal has no feathers} and so on.

\begin{defn}
	A \textbf{possibility} for an experimental domain $\edomain$ is a statement $x \in \tdomain$ that, when true, determines the truth value for all statements in the theoretical domain. Formally, $x \nequiv \contradiction$ and for each $\mathsf{s} \in \tdomain$, either $x \narrower \mathsf{s}$ or $x \ncomp \mathsf{s}$. The \textbf{possibilities} $X$ for $\edomain$ are the collection of all possibilities.
\end{defn}

The possibilities are all the different cases that can be distinguished experimentally given the verifiable statements of the domain. If we increase or otherwise change the set of verifiable statements (e.g. we learn how to test the DNA of animals) then the possibilities will change as well (e.g. the possible animal species are refined). We conclude this section with the following general result.

\begin{thrm}
	The possibilities $X$ for an experimental domain $\edomain$ have at most the cardinality of the continuum.
\end{thrm}

The proof is simply noting that each possibility can be labeled by the truth value of the countable basis of the experimental domain. We cannot have more possibilities than sequences of true/false and the set of all binary sequences has the cardinality of the continuum.

This means that we are never going to be able to experimentally distinguish between elements of greater cardinality. The set of topologically discontinuous functions from $\mathbb{R}$ to $\mathbb{R}$, for example, has greater cardinality and therefore it will never be associated with any physically distinguishable concept.\footnote{However, a discontinuous function with up to countably many discontinuities can be seen as the limit of a sequence of continuous functions and therefore may be used as a convenient approximation of a physical concept.} All the issues with large cardinals are not something science will ever be interested in. It does not matter what system we are describing, what experimental techniques we are using or how clever we are.

\section{Topologies and sigma-algebras}

\begin{table*}
	\centering
	\begin{tabular}{p{0.075\textwidth} p{0.275\textwidth} p{0.2\textwidth} p{0.3\textwidth}}
		& Statement relationship & & Set relationship  \\ 
		\hline 
		$\stmt_1 \AND \stmt_2$ & (Conjunction) & $U(\stmt_1) \cap U(\stmt_2)$ & (Intersection) \\ 
		$\stmt_1 \OR \stmt_2$ & (Disjunction) & $U(\stmt_1) \cup U(\stmt_2)$ & (Union) \\ 
		$\NOT \stmt$ & (Negation) & $U(\stmt)^C$ & (Complement) \\ 
		$\stmt_1 \equiv \stmt_2$ & (Equivalence) & $U(\stmt_1) = U(\stmt_2)$ & (Equality) \\ 
		$\stmt_1 \narrower \stmt_2$ & (Narrower than) & $U(\stmt_1) \subseteq U(\stmt_2)$ & (Subset) \\ 
		$\stmt_1 \broader \stmt_2$ & (Broader than) & $U(\stmt_1) \supseteq U(\stmt_2)$ & (Superset) \\ 
		$\stmt_1 \comp \stmt_2$ & (Compatibility) & $U(\stmt_1) \cap U(\stmt_2) \neq \emptyset$ & (Intersection not empty)
	\end{tabular} 
	\caption{Correspondence between statement operators and set operators.}\label{tab:statement_set}
\end{table*}

Now that we have introduced the basic mathematical structures for our general theory, we show their deep connection to other well established mathematical structures. We first note that each verifiable statement can be written as the disjunction of a set of possibilities.

\begin{defn}
	Let $\edomain$ be an experimental domain and $X$ its possibilities. We define the map $U : \edomain \rightarrow 2^X$ that for each statement $\obs \in \edomain$ returns the set of possibilities compatible with it. That is, $U(\obs)\equiv\{ x \in X \, | \, x \comp \obs\}$. We call $U(\obs)$ the \textbf{verifiable set} of possibilities associated with $\obs$.
\end{defn}

\begin{prop}
	A statement $\obs \in \edomain$ is the disjunction of the possibilities in its verifiable set $U(\obs)$. That is, $\obs=\bigOR\limits_{x \in U(\obs)} x$.
\end{prop}

The proof is a trivial application of the disjunctive normal form of Boolean algebra. In fact, each possibility can be written as a minterm of a basis (i.e. a conjunction where each basis element appears only once either negated or not). Any verifiable statement can be expressed in terms of the basis, and it is a result of Boolean algebra that each logical expression can be formulated as a disjunction of minterms (i.e. an OR of ANDs). Intuitively, it is the disjunction of all cases in which the statement is true.

Since every statement is a set of possibilities, we can re-express the statement relationships in terms of set relationships according to Table \ref{tab:statement_set}.

The closure of an experimental domain under finite conjunction and countable disjunction becomes closure under finite intersection and countable union. Since the basis is countable, countable union is equivalent to arbitrary union. In other words, the set of all verifiable sets is a topology.

\begin{thrm}
	Let $X$ be the set of possibilities for an experimental domain $\edomain$. $X$ has a natural topology given by the collection of all verifiable sets $\mathsf{T}_X=U(\edomain)$ that is Kolmogorov and second countable.
\end{thrm}

The topology is Kolmogorov (i.e. $T_0$) because given two possibilities, by their construction, there must be one element of the basis that is compatible with one but not the other. It is second countable because a basis of the experimental domain corresponds to a sub-basis of the topology. One can also show that the natural topology is Hausdorff if and only if all possibilities are approximately verifiable (i.e. each possibility is the limit of a sequence of verifiable statements). The hope is that we can find physically meaningful definitions for all relevant topological concepts.

In the same way, the theoretical domain corresponds to a $\sigma$-algebra on the possibilities.

\begin{defn}
	Let $\tdomain$ be a theoretical domain and $X$ its possibilities. We define the map $A : \tdomain \rightarrow 2^X$ that for each theoretical statement $\stmt \in \tdomain$ returns the set of possibilities compatible with it. That is, $A(\stmt)\equiv\{ x \in X \, | \, x \comp \stmt\}$. We call $A(\stmt)$ the \textbf{theoretical set} of possibilities associated with $\stmt$
\end{defn}

\begin{thrm}
	Let $X$ be the set of possibilities for a theoretical domain $\tdomain$. $X$ has a natural $\sigma$-algebra given by the collection of all theoretical sets $\Sigma_X=A(\tdomain)$.
\end{thrm}

The proof is again a simple mapping of statement operations to set operations. It can also be shown that the natural $\sigma$-algebra of the possibilities is the Borel algebra of their natural topology.

\section{Brief discussion and examples}

The connection we outlined creates a strong bridge between the standard mathematical structures used in physics and their meaning. Every theorem, every proof on those structures can now be given a direct physical meaning as well. And given the foundational nature of topologies and $\sigma$-algebras, we can imagine extending this framework to measure theory, differential geometry, symplectic geometry, Riemannian geometry, probability theory and so on.

To give an example of how it works, consider an experimental domain where the basis is composed of statements like \statement{this quantity is more than $x$ but less than $y$} where $x$ and $y$ are different values that can be arbitrarily close. For example, \statement{the distance between the earth and the moon is more than 384 but less than 385 thousand Km}. Each verifiable statement of the basis corresponds to an open interval of the real line, therefore we find a correspondence between arbitrary precision measurements and the standard topology on the reals, as this is the one generated by open intervals. Note, in fact, that this topology is second-countable and at least Kolmogorov.

The possibilities (e.g. \statement{the mass of the photon is precisely zero}) are not themselves verifiable since infinite precision measurements of a continuous quantity are not possible. Yet, their negation (i.e. \statement{the mass of the photon is not precisely zero}) could be verified in practice. This is expressed mathematically by the fact that singletons are closed sets. The theoretical domain, instead, corresponds to the standard Borel algebra, which includes closed and half-open sets but not all possible sets of integers, many of which cannot be characterized by a well formed formula.

Similarly, one can imagine verifiable statements for relationships (e.g. \statement{when the temperature of the mercury column is between 24 and 25 C, its height is between 24 and 25 mm}) and for statistical variables (e.g. \statement{if this coin is tossed enough times, the fraction of heads will be between 45\% and 55\%}). Ultimately the general theory will need to show precisely what mathematical structures map to experimental domains formed by statements of these types and under what assumptions. But the idea is that, once you have specified the verifiable statements and their logical relationships, you have already specified the experimental domain and nothing else needs to be added. In other words: \textbf{the points of the space and their mathematical structures are fully specified by what can be measured within the theory}.

\section{Conclusion}

Science is based on experimental verification. Experimental verification has its own logic. This logic imposes a mathematical structure which we characterized in terms of experimental domains (collections of verifiable statements), theoretical domains (collections of predictions for verifiable statements) and possibilities (the cases that can be distinguished experimentally). This mathematical structure leads naturally to topological spaces and $\sigma$-algebras, the foundation of many of the tools used in physics and science in general. Not only does this result clarify why those tools are so successful in science, it provides a direct physical meaning to those structures and a solid foundation upon which to create a general mathematical theory of experimental science.

In this light, what is most remarkable about this work is not its results, but that \emph{it can be done in the first place}. That there is a way to construct physical theories that forces us to spell out our physical assumptions, and that clarifies in a rigorous way what result is a consequence of what assumption. We can therefore analyze and compare new starting points with the discipline and thoroughness of modern mathematics instead of just rummaging through the bag of mathematical tools. But to do that, we have to create a space in the scientific community that is conducive to this kind of broad interdisciplinary work. We need generalists that can recognize how a detail in measure theory relates to information entropy or Hamiltonian mechanics. We need to take the mathematical structures created by mathematicians for their needs, break them apart and recombine them to find mathematical structures most suited and meaningful for science. Scientific knowledge has expanded quite considerably in the last century, maybe it is time for a moment of synthesis and consolidation.

We believe the development of a general mathematical theory of experimental science would be an important accomplishment for the scientific community and that it would be beneficial in ways we cannot yet even imagine. All we know is that a clearer understanding of our starting points cannot but help to provide insights into the current theories and suggest strategies to address long standing problems.

\section{Acknowledgments}

We thank Mark J. Greenfield for the thorough scrutiny of the mathematical framework and Mathew Timm for clarifying what can and cannot be defined within a formal mathematical framework. We thank Josh Hunt for the detailed review of the philosophical underpinning of this work and Gordon Belot, Laura Ruetsche and David J. Baker for philosophical discussions related to this work. We would also like to acknowledge John Mayer, Kai Sun, Jens Zorn and others for their interest and support for this project.

\bibliography{bibliography}

\end{document}